%% file: main.tex
\documentclass[10pt,twocolumn,letterpaper]{article}

\usepackage{cvpr}              

\input{preamble}

\definecolor{cvprblue}{rgb}{0.21,0.49,0.74}
\usepackage[pagebackref,breaklinks,colorlinks,citecolor=cvprblue]{hyperref}
\usepackage{kotex}
\usepackage{lipsum}
\usepackage{comment}
\usepackage{algorithm}
\usepackage{algpseudocode}
\usepackage{courier}
\usepackage{stackengine}
\usepackage{float}
\usepackage{booktabs}       %

\def\paperID{5733} 
\def\confName{CVPR}
\def\confYear{2024}

\title{Neutral Editing Framework for Diffusion-based Video Editing}

\author{Sunjae Yoon \hspace{0.5cm} Gwanhyeong Koo \hspace{0.5cm} Ji Woo Hong \hspace{0.5cm} Chang D. Yoo\\
Korea Advanced Institute of Science and Technology (KAIST)\\
{\tt\small \{sunjae.yoon,kookie,jiwoohong93,cd\_yoo\}@kaist.ac.kr}
}

\begin{document}
\twocolumn[{
\maketitle
\begin{center}
   \includegraphics[width=1.0\textwidth]{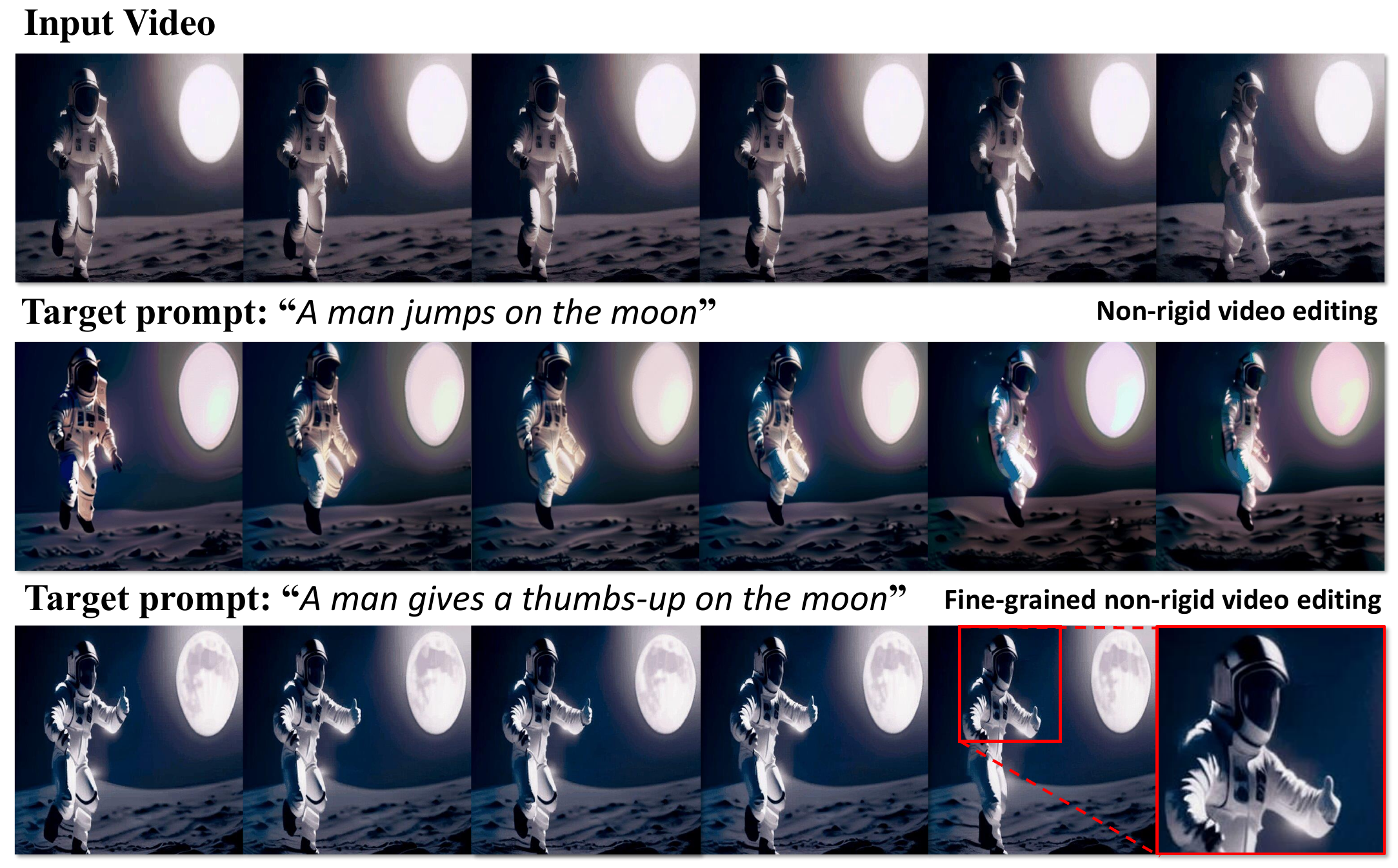}
   \captionof{figure}{Neutral editing framework enables the diffusion-based editing models to perform various text-based non-rigid editing such as motion variation of objects spanning from fine-grained variations to large dynamic variations while preserving fidelity to the input video.}
\label{fig:teaser}
\end{center}
}]

\input{sec/0_abstract}    
\input{sec/1_intro}

\input{sec/2_Related_Works}
\input{sec/3_Preliminary}
\input{sec/4_Model}
\input{sec/5_Experiment}
\newpage
{
    \small
    \bibliographystyle{ieeenat_fullname}
    \bibliography{main}
}

\input{sec/X_suppl}
\end{document}

%% file: preamble.tex
\usepackage[dvipsnames]{xcolor}


%% file: sec/0_abstract.tex

\begin{abstract}
    Text-conditioned image editing has succeeded in various types of editing based on a diffusion framework. Unfortunately, this success did not carry over to a video, which continues to be challenging. Existing video editing systems are still limited to rigid-type editing such as style transfer and object overlay. To this end, this paper proposes Neutral Editing (NeuEdit) framework to enable complex non-rigid editing by changing the motion of a person/object in a video, which has never been attempted before. NeuEdit introduces a concept of `neutralization' that enhances a tuning-editing process of diffusion-based editing systems in a model-agnostic manner by leveraging input video and text without any other auxiliary aids (e.g., visual masks, video captions). Extensive experiments on numerous videos demonstrate adaptability and effectiveness of the NeuEdit framework. The website of our work is available here: \href{https://neuedit.github.io}{https://neuedit.github.io/}
\end{abstract}

%% file: sec/1_intro.tex
\begin{figure*}[t!]
\centering
    \includegraphics[width=1.0\textwidth]{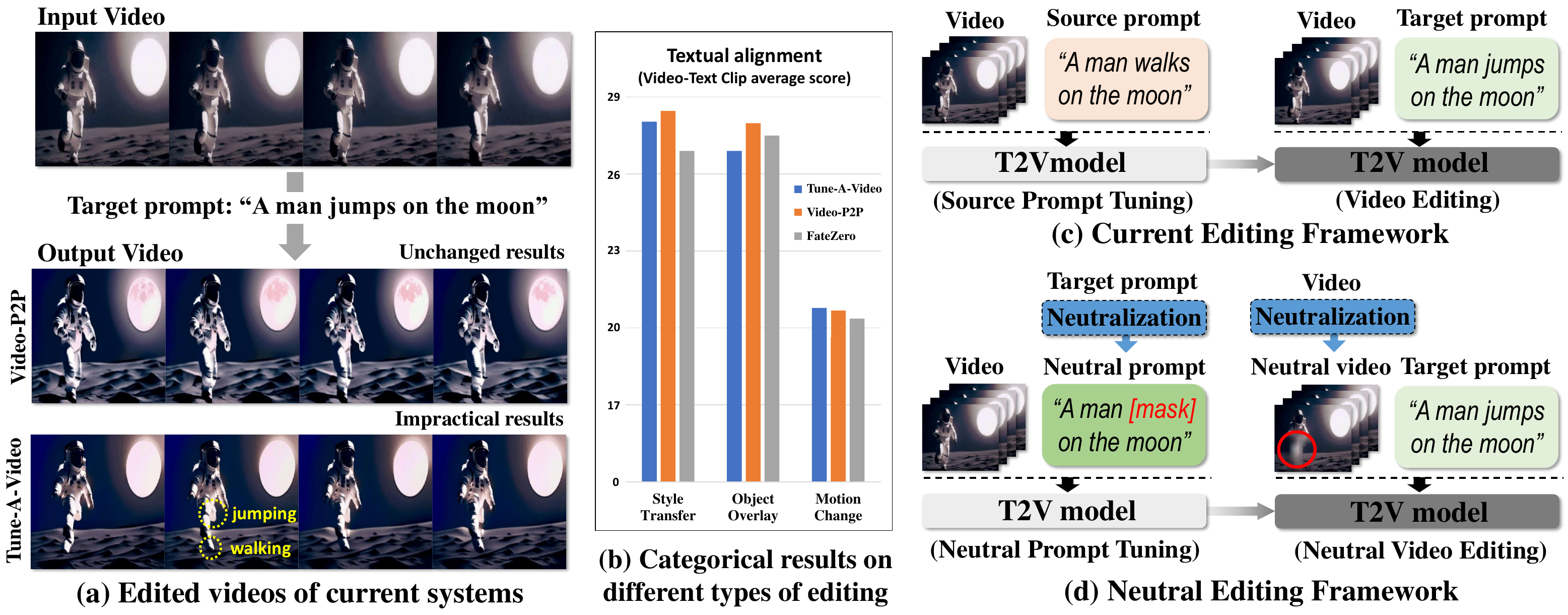}
   \caption{(a) Edited videos of current systems \cite{wu2022tune,liu2023video} based on target prompt in terms of motion change. (b) Categorical analysis of different types of editing on videos of DAVIS \cite{pont20172017}. Illustration of (c) current editing framework and (d) proposed neutral editing framework.}
\label{fig:introduction_1}
\end{figure*}
\section{Introduction}
\label{sec:intro}
%
The recent success of generative frameworks \citep{kingma2013auto,ho2020denoising,creswell2018generative} and large-scale models \citep{devlin2018bert,radford2018improving,radford2021learning} provide surreal outputs surpassing the boundaries of human capabilities.
%
%
The diffusion models \citep{sohl2015deep,song2020score,dhariwal2021diffusion} lay the foundation for such innovative advancements, bridging a diverse range of large-scale generative models \citep{song2020denoising,rombach2022high}.
%
To be specific, diffusion-based text-to-image (T2I) models \citep{rombach2022high,nichol2021glide} synthesize natural images of high fidelity and further edit \citep{ruiz2022dreambooth,kawar2022imagic} them by modifying specific attributes corresponding to the input text.
%
Expanding the work in image, diffusion-based text-to-video (T2V) models \citep{bar2022text2live,wu2022tune} also have been considered.
%
Due to insufficient training resources about videos, in early work, significant technical contributions \citep{singer2022make,hong2022cogvideo} have been made to transfer the knowledge of pre-trained T2I models into the T2V models.
Currently, researchers are striving to refine this text-based video generation into a more controlled and fine-grained approach by modifying specific attributes in a video corresponding to users' requirements from a text, ultimately performing text-based video editing.
%
%
%

%
%
%
In a formal definition of text-based video editing, as shown in Figure \ref{fig:introduction_1} (a), systems are given an input video and a target textual prompt describing desired modifications in the video such that they produce an edited video that conforms to the target prompt.
To achieve this, editing systems largely perform two sequential processes: (1) video tuning and (2) video editing.
In video tuning, the editing system is trained to generate the input video and to comprehend the contextual meaning of the video.
In video editing, the system generates the variants of the input video that conform to the meaning of the target prompt.
To provide necessary attributes for editing, pre-trained vision-language models \citep{radford2021learning,rombach2022high} also have to be integrated into the system.

Despite recent advancements in video editing systems, their capabilities are still restricted to rigid modifications within the realm of inpainting such as style transfer and object overlay.
%
To be specific, in Figure \ref{fig:introduction_1} (a), for a given target prompt (\eg ``A man jumps on the moon'') requiring non-rigid modifications by changing a motion of an object, current systems do not conform to the target prompt and return the original input video under over-fidelity.
%
Otherwise, they often show impractical results by mixing up the original content (\ie walking) and targeted content (\ie jumping).
In Figure \ref{fig:introduction_1} (b), our categorical analysis of textual alignment with video according to different types of editing (\ie style transfer, object overlay, motion change) demonstrates that current systems are facing difficulties in changing a motion in a video.
Therefore the results of complex non-rigid editing are still unsatisfactory.
%

%
One of the reasons for the unsatisfactory editing is rooted in a conventional tuning and editing process within diffusion-based editing frameworks.
As illustrated in Figure \ref{fig:introduction_1} (c), current editing frameworks require additional input caption about the video (\ie source prompt) in the tuning process. 
%
%
After tuning, the model edits the video based on a target prompt.
However, employing a source prompt leads to a functionally unnecessary tuning of content (\eg `walking') in the video, which is unrelated to the intended editing (\eg `jumping') and results in suboptimal editing. 
Furthermore, the outcomes are vulnerable to the variants of source prompts.
Therefore, frameworks employing source prompt are inadequate for effective text-based video editing.

To this end, we propose Neutral Editing (NeuEdit) framework that performs effective video editing including non-rigid editing to a video with only a target prompt.
As shown in Figure \ref{fig:introduction_1} (d), the NeuEdit framework introduces a novel concept of \textit{neutralization}, which enables current editing systems to conduct (1) neutral prompt tuning and (2) neutral video editing.
The neutral prompt tuning refers to tuning a model based on a neutral prompt.
This prompt (\eg ``A man [mask] on the moon") is a text that reduces\footnote{Masking is an intuitive approach for reducing the factors. See also other approaches in Method.} factors (\eg ``jumps") contributing to editing from a target prompt, allowing a model to tune a video without relying on a source prompt. 
Furthermore, the target prompt holds effective differences from this neutral prompt by the factors related to editing.
To implement a neutral prompt, we first introduce a neutralization which refers to disentangling a factor related to editing in the input.
After tuning with the neutral prompt, the T2V model performs video editing.
Our studies found that current models struggle with non-rigid editing, primarily due to constraints imposed by the original content in the input video (\eg Figure \ref{fig:introduction_1} (a)).
To address this, we also construct a neutral video by applying neutralization which reduces the influence of original content in a region of the video to be edited, such that it amplifies the possibility of non-rigid editing.
NeuEdit can be applied to diffusion-based editing systems in a model-agnostic manner, enhancing various editing including object motion change.
%
Extensive experiments validate its adaptability and visual effectiveness.
%
%
%
%
%
%
%
%
%


%% file: sec/2_Related_Works.tex
\section{Related Works}
\label{sec:formatting}
\subsection{Diffusion-based generative models}
Deep diffusion models \citep{ho2020denoising,song2020denoising} exhibit significant capability by outperforming the prior best qualities of generative adversarial networks \citep{goodfellow2020generative}.
%
%
%
Applying diffusion to a text-to-image (T2I) generation, significant advancements have been observed in image generation, where diffusion-based T2I models \citep{ramesh2022hierarchical,saharia2022photorealistic} produce high-fidelity images from textual inputs. 
Recently, the T2I models have expanded their visual generative capabilities into the domain of videos to perform text-to-video (T2V) generation.
%
%
%
%
Earlier studies \citep{ho2022video,wu2022nuwa,hong2022cogvideo} in T2V generation modified the T2I model by introducing a temporal axis for video data, thus transferring pre-trained knowledge from the T2I model.
%
To enhance temporal consistency in generated frames, temporal attentions \citep{ho2022imagen,singer2022make} are also designed.
Recently, these diffusion-based models have been successful in various generative works including inpainting and super-resolution \citep{saharia2022image,lugmayr2022repaint}. 
Among them, visual editing emerges as a new challenge to perform controlling and reasoning about selective synthesizing, which is discussed in detail below.
%
%

\subsection{Image and video editing }
%
%
%
Text-based image editing aims to edit a given image based on text descriptions.
To perform this, DiffusionCLIP \cite{kim2022diffusionclip} first proposes a tuning-editing framework of diffusion model based on CLIP embedding and Prompt-To-Prompt \citep{hertz2022prompt} proposes weight blending to perform effective rigid editing in this framework.
%
%
%
%
%
For efficient editing, InstructPix2Pix \citep{brooks2022instructpix2pix} design zero-shot edits without fine-tuning.
%
%
Similar to work in the image, video editing also has expanded.
%
%
Especially to keep temporal consistency, several technical solutions are introduced such as layered editing \cite{chai2023stablevideo} and attention control \cite{liu2023video}.
%
%
However, current video editing is limited to rigid types of editing and still challenging to dynamic motion change.
Thus, NeuEdit first performs complex non-rigid editing in a video based on a text.

%% file: sec/3_Preliminary.tex
\section{Preliminaries}
\label{sec:prelim}
\subsection{Denoising diffusion probabilistic models}
\label{gen_inst}
Denoising diffusion probabilistic models (DDPMs) \cite{ho2020denoising} are parameterized Markov chains to reconstruct a sequence of data $\{x_{1}$,$\cdots$, $x_{T}\}$.
Given raw data $x_{0}$, the Markov transition gradually adds Gaussian noise upto $x_{T}$ using $q(x_{t}|x_{t-1}) = \mathcal{N}(x_{t};\sqrt{\alpha_{t}}x_{t-1},(1-\alpha_{t})I)$ under pre-defined schedule $\alpha_{t}$ following $t=1,\cdots,T$.
This process is referred to as a \textit{forward process} of the diffusion model.
%
%
In the \textit{reverse process}, the diffusion model approximates the $q(x_{t-1}|x_{t})$ using trainable Gaussian
transitions $p_{\theta}(x_{t-1}|x_{t}) = \mathcal{N}(x_{t-1};\mu_{\theta}(x_{t},t),\sigma_{\theta}(x_{t},t))$ starting at normal distribution $p(x_{T}) = \mathcal{N}(x_{T};0,I)$.
The training objective is to maximize log-likelihood $log(p_{\theta}(x_{
0}))$, where we can also apply variational inference by maximizing the variational lower bound of this.
This makes a closed-form of KL divergence\footnote{See the detailed proof in Appendix D.} between the distributions of $p_{\theta}$ and $q$ while optimizing the parameter $\theta$.
The beauty of DDPM is that this process can be summarized as denoising network $\epsilon_{\theta}(x_{t}, t)$ for predicting noise $\epsilon \sim \mathcal{N}(0,I)$ as given below:
%
\begin{equation}
\begin{aligned}
\mathbb{E}_{x,\epsilon \sim \mathcal{N}(0,1),t \sim \mathcal{U}\{1,T\}}[||\epsilon - \epsilon_{\theta}(x_{t},t)||_{2}^{2}].
\end{aligned}
\end{equation}
To keep the robustness in all steps, $t$ is sampled from the discrete uniform distribution $\mathcal{U}\{1,T\}$.
Based on trained $\epsilon_{\theta}$, denoising is performed, where denoising diffusion implicit model (DDIM) \citep{song2020denoising} has been a popular choice for the denoising by a small number of sampling steps as below:
\begin{equation}
x_{t-1} = \sqrt{\frac{\alpha_{t-1}}{\alpha_{t}}}x_{t} + \left( \sqrt{\frac{1-\alpha_{t-1}}{\alpha_{t-1}}} - \sqrt{\frac{1-\alpha_{t}}{\alpha_{t}}} \right)\cdot \epsilon_{\theta}.
\label{eq:ddimeq}
\end{equation}
\label{eq:ddim}
%
%
%
%
\subsection{Text-guided diffusion model}
\label{sec:3.2}
The text-guided diffusion model is a DDPM that restores the output data $x_{0}$ from random noise with a guided condition of a text prompt $\mathcal{T}$.
Thus, the training objective is also formulated with this condition under latent space to interact with textual modality as $\mathbb{E}_{z,\epsilon,t}[||\epsilon - \epsilon_{\theta}(z_{t},t,\mathbf{c})||_{2}^{2}]$, where $z_{t} = E(x_{t})$ is a latent noise encoding (\eg VQ-VAE \cite{van2017neural}) and $\mathbf{c}=\psi(\mathcal{T})$ is conditional textual embedding (\eg CLIP \cite{radford2021learning}).
%
%
%
%
In video editing, $z_{t}$ is a latent encoding of video data, and $\epsilon_{\theta}$ can be pre-trained video diffusion networks.

%% file: sec/4_Model.tex
\begin{figure*}[t!]
\centering
    \includegraphics[width=1.0\linewidth]{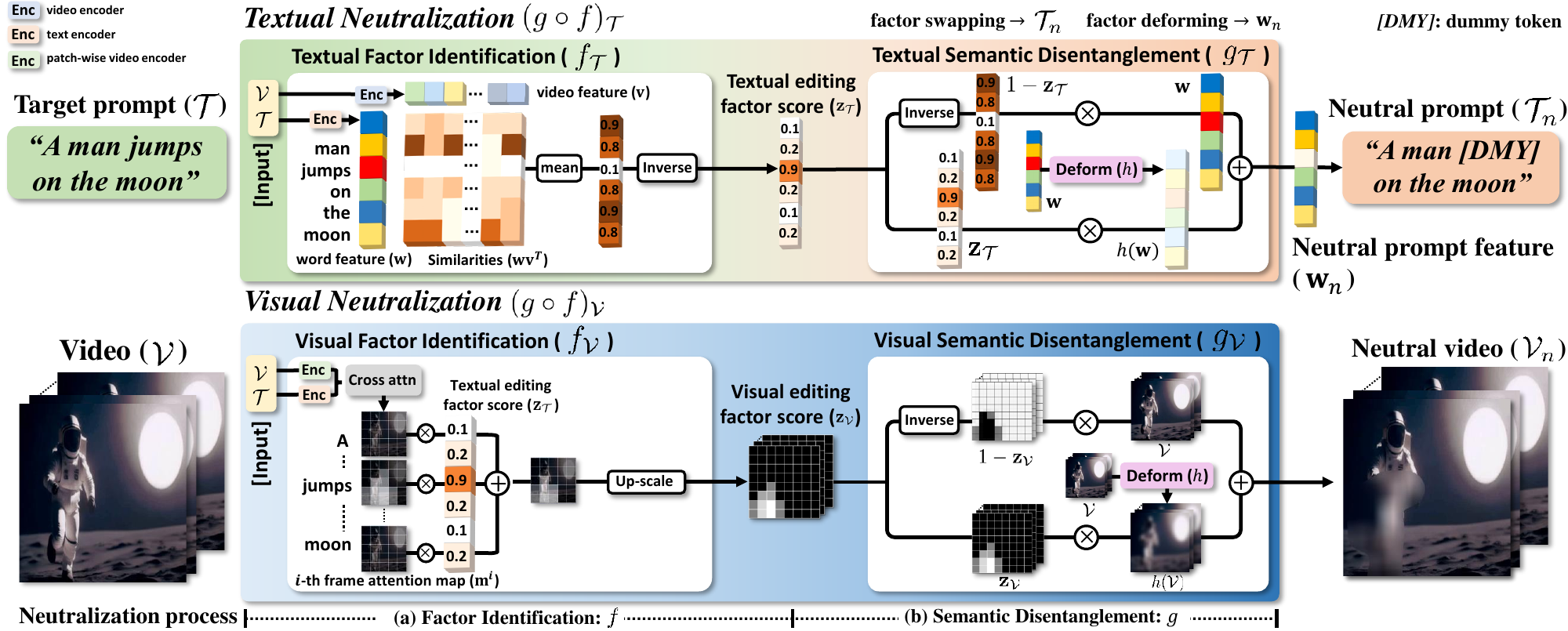}
   \caption{Illustration of neutralization composed of (a) factor identification $f$ and (b) semantic disentanglement $g$. The $f$ localizes editing factors in video and text and the $g$ produces neutralized video and text via semantically reducing the editing factors in the video and text. The textual neutralization $(g \circ f)_{\mathcal{T}}$ and visual neutralization $(g \circ f)_{\mathcal{V}}$ are the applications of $f$ and $g$ for performing the neutralization.}
\label{fig:neutralizing}
\end{figure*}
\begin{figure}[t]
\centering
    \includegraphics[width=1.0\linewidth]{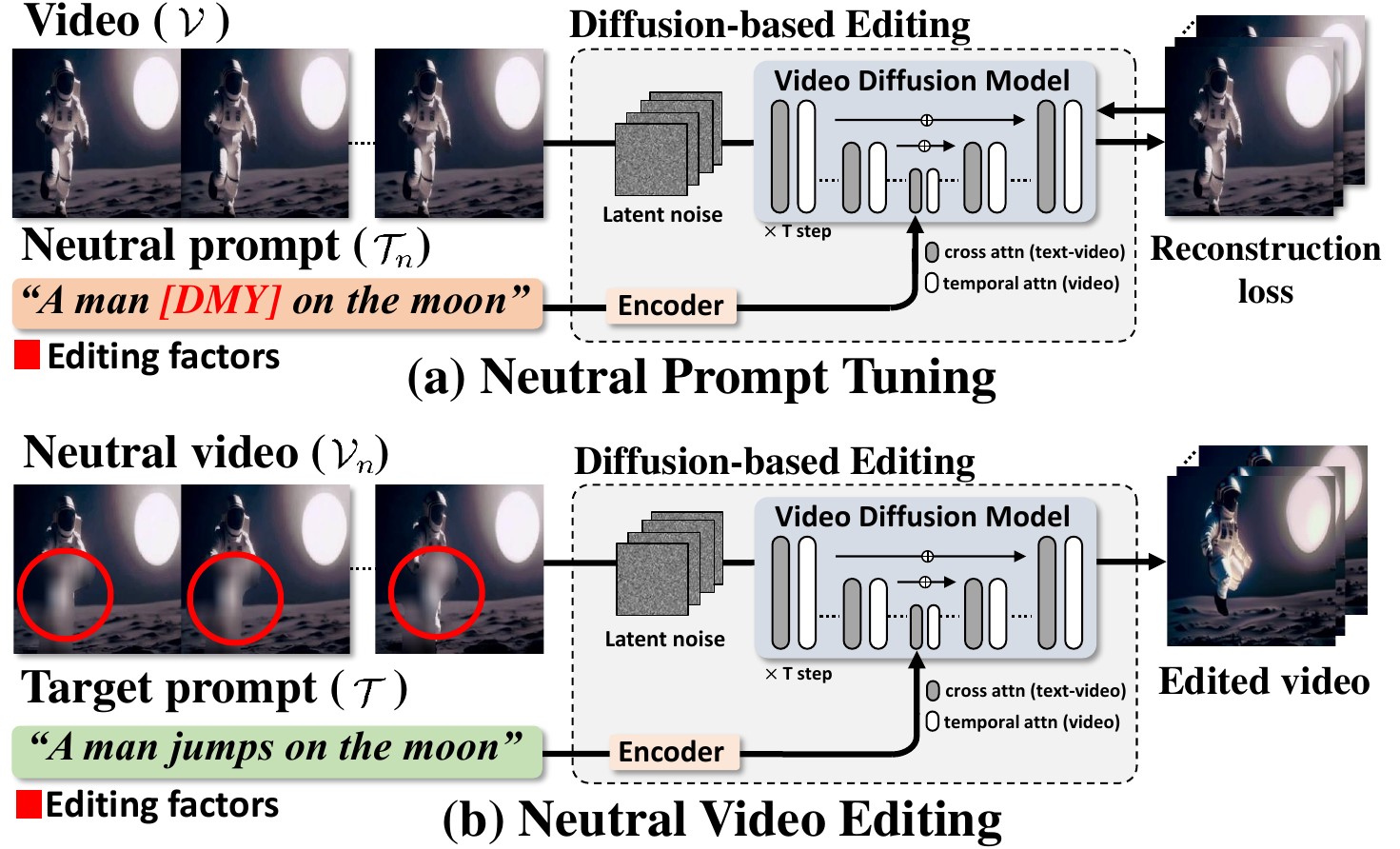}
   \caption{Illustration of Neutral Editing framework composed of (a) neutral prompt tuning and (b) neutral video editing. The neutral prompt is a text that removes factors contributing to editing from the target prompt. The neutral video is a video that reduces factors related to editing enabling the systems to perform effective editing including non-rigid editing without auxiliary inputs.}
\label{fig:model}
\end{figure}
\section{Neutral Editing Framework}
The Neutral Editing (NeuEdit) framework aims to enhance existing diffusion-based video editing systems, enabling more effective non-rigid modifications in a model-agnostic approach using only an input video and a target prompt.
To achieve this, as shown in Figure \ref{fig:model} (a), NeuEdit introduces the novel concept of neutralization which enables current editing systems to conduct (1) neutral prompt tuning and (2) neutral video editing.
The neutral prompt tuning refers to model tuning based on a neutral prompt. 
This neutral prompt is a text that reduces factors (\eg specific words or features) contributing to editing from a target prompt, effectively resolving the issue of spurious reliance on additional source prompt of current editing systems.
Henceforth, the target prompt keeps effective differences from the neutral prompt by the factors related to editing.
%
%
After tuning, the systems edit a video based on the target prompt. 
However, our studies found that original content within the editing region in the video imposes constraints on non-rigid edits.
Thus we construct a neutral video that sensibly reduces the influence of original content in a region of video to be edited, amplifying the possibilities of non-rigid editing.
The neutral prompt and neutral video are constructed by following our proposed neutralization operation.
%
%
\subsection{Neutralization}
Neutralization aims to disentangle factors contributing to editing in the input modality (\ie video, text). 
To perform this, as shown in Figure \ref{fig:neutralizing}, it comprises two sequential processes: (1) factor identification $f$ and (2) semantic disentanglement $g$.
To provide a formal definition of the neutralization, it takes inputs of target prompt $\mathcal{T}$ and video $\mathcal{V}$ and produces neutral prompt $\mathcal{T}_{n}$ and neutral video $\mathcal{V}_{n}$ as below: 
\begin{equation}
\mathcal{T}_{n}, \mathcal{V}_{n} = (g \circ f)(\mathcal{T},\mathcal{V}),
\end{equation}
where $f$ is factor identification which localizes the editing factors within each modality, and $g$ is semantic disentanglement that produces the modality that semantically reduces the meaning of the identified editing factors.
The implementations of $f$ and $g$ are specified depending on the target (\ie text or video) of neutralization, referred to as textual neutralization and visual neutralization in the following.
%
\paragraph{Textual Neutralization}
Textual neutralization $(g \circ f)_{\mathcal{T}}$ at the top of Figure \ref{fig:neutralizing} is an application of the neutralization to the target prompt, which obtains neutral prompt $\mathcal{T}_{n}$ from target prompt $\mathcal{T}$ and video $\mathcal{V}$.
To implement the factor identification $f_{\mathcal{T}}$ and the semantic disentanglement $g_{\mathcal{T}}$ in the textual neutralization, we first define the editing factors in the target prompt as `words' contributing to editing.
Thus, the $f_{\mathcal{T}}$ aims to localize these words in the target prompt $\mathcal{T}$.
Since the target prompt is a description of desired modifications in the current video, the prompt and the video exhibit semantic misalignment due to the words associated with the modifications (\ie editing factors).
Intrigued by this observation, we measure all words in $\mathcal{T}$ based on their cosine similarities with $\mathcal{V}$ to localize the words exhibiting low similarities as the editing factors.
Therefore, we define a textual factor identification as $f_{\mathcal{T}} (\mathcal{T},\mathcal{V})$ to produce scores about localization of editing factor in the target prompt based on their similarity scores as given below:
\begin{equation}
f_{\mathcal{T}}(\mathcal{T},\mathcal{V}) = 1 - \textrm{mean}(\mathbf{w}\mathbf{v}^{\top}) \in \mathbb{R}^{M},
\end{equation}
where the $\mathbf{w} = \psi_{T}(\mathcal{T}) \in \mathbb{R}^{M \times d}$ is $d$-dimensional normalized word features of target prompt. The $M$ is the number of words and $\psi_{T}$ is the CLIP \cite{radford2021learning} text encoder. The $\mathbf{v} = \psi_{I}(\mathcal{V}) \in \mathbb{R}^{L \times d}$ is $d$-dimensional video features with frame length $L$, where $\psi_{I}$ is the CLIP image encoder.
$\textrm{mean}(\cdot)$ is a mean-pooling along frame axis.
As editing factors have low similarity scores with video, we inverse the scores by subtracting them from one, finally producing the textual editing factor score denoted as $\mathbf{z}_{\mathcal{T}} = f_{\mathcal{T}}(\mathcal{T},\mathcal{V})$.
This score $\mathbf{z}_{\mathcal{T}}$ is utilized for the following textual semantic disentanglement.

The textual semantic disentanglement $g_{\mathcal{T}}$ aims to build a neutral prompt $\mathcal{T}_{n}$ from a target prompt $\mathcal{T}$ by semantically reducing editing factors in the $\mathcal{T}$.
To perform this, $g_{\mathcal{T}}$ employs the editing factor score $\mathbf{z}_{\mathcal{T}}$ to identify the editing factors and reduce their meaning by disentangling them.
To be specific, we present two technical contributions about the disentangling methods: (1) factor swapping and (2) factor deforming.
The factor swapping is to swap the identified editing factors with other words. 
To the scores above a specific value $s$ (\eg 0.7) on $\mathbf{z}_{\mathcal{T}}$ (\eg [0.1, 0.9, $\cdots$, 0.2]), their corresponding words are decided as editing factors, denoting them as $W_{\mathbf{z}_{\mathcal{T}}>s}$ in word space, where $\mathbf{z}_{\mathcal{T}}>s$ is the indices of words in target prompt $\mathcal{T}$ of a higher score than $s$. 
Thus, the $W_{\mathbf{z}_{\mathcal{T}}>s}$ are swapped with other word tokens, where to mitigate a semantic intervention by swapping tokens, we define a dummy token as $<$\textrm{DMY}$>$ for the swapping.
As a result, the textual semantic disentanglement with factor swapping ultimately produces a neutral prompt $g_{\mathcal{T}}(z_{\mathcal{T}}) = \mathcal{T}_{n} = [W_{1},\cdots,W_{i},\cdots,W_{M}]$ including $W_{\mathbf{z}_{\mathcal{T}} > s} =$ $<$\textrm{DMY}$>$, where $W_{i}$ denotes corresponding $i$-th word in the target prompt $\mathcal{T}$.
Although the $\mathcal{T}_{n}$ with factor swapping maintains a distinct difference\footnote{Experimental studies in Appendix F provide optimal region of the $s$.} with the $\mathcal{T}$ by the editing factors $W_{\mathbf{z}_{\mathcal{T}} > s}$, it relies on a heuristic manner in selecting the editing factor and is difficult to distinguish the difference among the factors.
To this end, we further devised a feature-level disentanglement referred to as factor deforming.
To be specific, it first disentangles the target prompt features $\mathbf{w} \in \mathbb{R}^{M \times d}$ into a format of linear combination using factor score $\mathbf{z}_{\mathcal{T}} \in \mathbb{R}^{M}$ as $\mathbf{w} = \mathbf{z}_{\mathcal{T}} \circ \mathbf{w} + (1 - \mathbf{z}_{\mathcal{T}}) \circ \mathbf{w}$, where $\circ$\footnote{Here, $\circ$ is different from composite operation in $f \circ g$} is element-wise multiplication with broadcasting.
After that, we deform the features attended by $\mathbf{z}_{\mathcal{T}}$ using deformable operation $h(\cdot)$ as:
%
%
\begin{equation}
\mathbf{w}_{n} = \mathbf{z}_{\mathcal{T}} \circ h(\mathbf{w}) + (1 - \mathbf{z}_{\mathcal{T}}) \circ \mathbf{w}.
\end{equation}
This format selectively deforms the text features concerning the editing factor while keeping disparities among features of editing factors.
%
We apply feature down-scaling for the deforming\footnote{Appendix F also provides other factor deforming methods} as $h(\mathbf{w}) = \alpha \times \mathbf{w}$ with scaler $0 \leq \alpha < 1$.
For $\alpha=0$, all deformed features become identical, working similarly to the factor swapping.
Finally, we define a neutral prompt features $g_{\mathcal{T}}(\mathbf{z}_{\mathcal{T}})=\mathbf{w}_{n} \in \mathbb{R}^{M \times d}$, which is utilized in tuning process of NeuEdit framework.
\paragraph{Visual Neutralization}
Visual neutralization $(g \circ f)_{\mathcal{V}}$ at the bottom of Figure \ref{fig:neutralizing} is another application of neutralization to the input video, which produces neutral video $\mathcal{V}_{n}$ from target prompt $\mathcal{T}$ and video $\mathcal{V}$.
Neutral video improves the effectiveness of editing by reducing the influence of the original content in a region to be edited.
To construct a factor identification $f_{\mathcal{V}}$ and semantic disentanglement $g_{\mathcal{V}}$ in the visual neutralization, we also define the editing factors in the video as `pixels' contributing to editing.
Thus the $f_{\mathcal{V}}$ aims to localize these pixels in the input video $\mathcal{V}$.
As editing models' multi-modal interaction modules (\ie cross-attention)\footnote{Video editing is based on pre-trained knowledge (\eg Stable Diffusion \cite{rombach2022high}) with multi-modal attention to provide required modifications.} contain information about the interactions between texts and frames, we employ this information to build the $f_{\mathcal{V}}$.
We first embed each $i$-th frame of video into patch-wise features as $\mathbf{p}^{i} \in \mathbb{R}^{(W_{p} \times H_{p}) \times d}$ and perform cross attention with a target prompt features $\mathbf{w} \in \mathbb{R}^{M \times d}$ to get cross attention maps as $\mathbf{m}^{i} \in \mathbb{R}^{(W_{p} \times H_{p}) \times M}$, where ($W_{p} \times H_{p}$) is the number patches in the frame and $M$ is the number of words in the target prompt.
Among the attention maps $\mathbf{m}^{i}$, we highlight the maps related to the editing using textual editing factor score $\mathbf{z}_{\mathcal{T}} \in \mathbb{R}^{M \times 1}$ as (Please see also illustration in the bottom of Figure \ref{fig:neutralizing} for clear understanding.):
\begin{equation}
\mathbf{z}_{\mathcal{V}}^{i} = \mathbf{m}^{i}  \mathbf{z}_{\mathcal{T}} \in \mathbb{R}^{W_{p} \times H_{p}},
\end{equation}
where $\mathbf{z}_{\mathcal{V}}^{i}$ is the $i$-th frame visual editing factor score.
After restoring $\mathbf{z}_{\mathcal{V}}^{i}$ up to the original frame scale ($W \times H$) and aggregating all frames, we finally define the visual editing identification as $f_\mathcal{V}(\mathcal{T},\mathcal{V}) = \mathbf{z}_{\mathcal{V}} \in \mathbb{R}^{L \times (W \times H)}$, where $L$ is the number of video frames.
%
%
The visual factor score $\mathbf{z}_{\mathcal{V}}$ is utilized for the following visual semantic disentanglement.

The visual semantic disentanglement $g_{\mathcal{V}}$ aims to build neutral video $\mathcal{V}_{n}$ from input video $\mathcal{V}$ by reducing the meaning of editing factors in the video.
To this, based on visual editing factor score $\mathbf{z}_{\mathcal{V}}$, the $g_{\mathcal{V}}$ identifies factors contributing to editing at a pixel level and semantically reduces them.
Similar to textual semantic disentanglement $g_{\mathcal{T}}$, we apply a factor deforming by separating video pixels into two groups of pixels and deforming a group related to editing as below:
\begin{equation}
\mathcal{V}_{n} = \mathbf{z}_{\mathcal{V}} \circ h(\mathcal{V}) + (1 - \mathbf{z}_{\mathcal{V}}) \circ \mathcal{V},
\end{equation}
where we applied Gaussian blurring to deform the video concerned about editing factors as $h(\mathcal{V}) = \mathcal{V}*G$ with Gaussian kernel $G(x,y) = \frac{1}{2\pi \sigma^{2}}e^{-(x^2+y^2)/2\sigma^2}$. Therefore the visual semantic disentanglement is summarized as $g_{\mathcal{V}}(\mathbf{z}_{\mathcal{V}})=\mathcal{V}_{n}$.
The $\mathcal{V}_{n}$ is used for editing instead of the $\mathcal{V}$, alleviating restrictions imposed by the original content and facilitating dynamic variations within the editing areas.
%
\subsection{Plug-and-play NeuEdit framework}
We integrate the neutral prompt $\mathcal{T}_{n}$ and neutral video $\mathcal{V}_{n}$ into diffusion-based video editing system.
The editing system includes two processes: (1) video tuning and (2) video editing, where the neutral prompt is introduced in the tuning process as a guided condition features $\mathbf{w}_{n} = \psi_{T}(\mathcal{T}_{n})$ for the training objective of a text-guided diffusion model (\ie refer details in Section \ref{eq:ddim}) as given below:
\begin{equation}
\mathbb{E}_{z,\epsilon ,t}[||\epsilon - \epsilon_{\theta}(z_{t},t,\mathbf{w}_{n})||_{2}^{2}],
\end{equation}
%
%
%
%
%
%
%
After tuning with neutral prompt, the model performs editing by denoising an initial latent noise with the input condition of target prompt $\mathcal{T}$, producing edited video $\mathcal{V}_{\textrm{edit}}$ as:
\begin{equation}
\mathcal{V}_{\textrm{edit}} = \mathrm{\textrm{Denoise}}(f_{init}(\mathcal{V}_{n}),\mathcal{T}),
\end{equation}
where $\textrm{Denoise}(\cdot,\cdot)$ is the reverse process of diffusion model by gradual denoising under the sequential process using Equation \ref{eq:ddimeq} and $f_{init}$ is initial latent noise encoding such as DDIM inversion\footnote{Appendix E gives details of DDIM inversion and gradual denoising.} for enhanced reconstruction based on input video, where we provide neutral video $\mathcal{V}_{n}$ instead of $\mathcal{V}$ to improve dynamic modifications in editing region.
%
%
%
\label{headings}

%% file: sec/5_Experiment.tex
\begin{figure*}[t!]
\centering
    \includegraphics[width=1.0\textwidth]{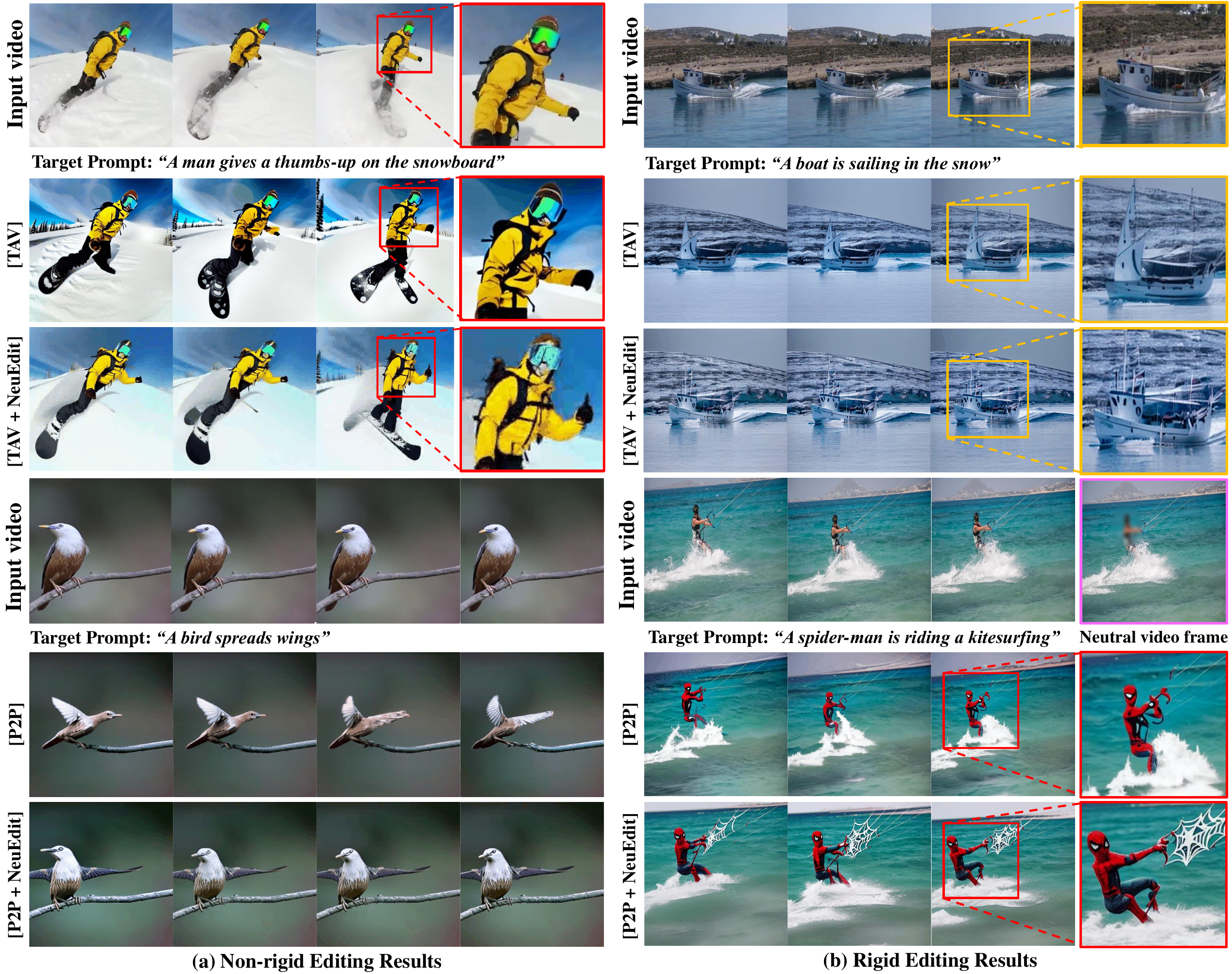}
   \caption{Qualitative result about applying NeuEdit framework on recent editing systems according to (a) non-rigid editing (\ie motion variation) and (b) rigid editing (\ie style transfer, object overlay). (TAV: Tune-A-Video \cite{wu2022tune}, P2P: Video-P2P \cite{liu2023video}).}
\label{fig:qual}
\end{figure*}
\section{Experiment}
\subsection{Experimental Settings}
\paragraph{Implementation Details.} For textual factor identification $f_\mathcal{T}$, CLIP model (ViT-L/14) \cite{radford2021learning} is used for text and image features.
The attention map for visual neutralization is based on the cross attention module of video diffusion model \cite{wu2022tune} (\ie Stable Diffusion v1.5).
%
%
The experimental settings are $W = H =512, W_{p}= H_{p} = 16$ on NVIDIA A100 GPU.
%
%
%
%
%
%
\paragraph{Dataset and Baselines.}
We validate videos on DAVIS \citep{pont20172017} and LOVEU-TGVE \cite{wu2023cvpr}, which are video editing challenge dataset\footnote{https://sites.google.com/view/loveucvpr23/track4} comprising 32 to 128 frames of each. 
%
NeuEdit framework is validated about non-rigid/rigid editing on recent editing systems including Tune-A-video, Video-P2P, FateZero \cite{qi2023fatezero} on their public codes.
%
%
\subsection{Evaluation Metric}
We validate editing results based on four assessments: (1) textual alignment, (2) fidelity to input video, (3) frame consistency, and (4) human preference.
The textual alignment measures the semantic alignment between a target prompt and an edited video using the CLIP score and PickScore \cite{kirstain2023pick}.
The PickScore approximates human preferences by a large-scale trained model.
The fidelity measures the preservation of original content in the unedited region\footnote{Detailed explanations of capturing unedited region are in Appendix C.} using peak signal-to-noise ratio (PSNR), learned perceptual image patch similarity (LPIPS), and structural similarity index measure (SSIM).
The frame consistency measures image CLIP scores between sequential frames and measures fréchet video distance (FVD) to evaluate the naturalness of videos.
For the human evaluation, we investigate the preferences of edited videos according to the target prompt between the editing models and the models with NeuEdit.
\subsection{Experimental Results}
\paragraph{Qualitative Comparisons.}
Figure \ref{fig:qual} shows the qualitative results of recent editing systems \cite{wu2022tune,liu2023video} with the proposed neutral editing framework. (See also qualitative results in Appendix G).
To validate the qualitative effectiveness of neutralization, we perform case studies in terms of two types of editing: (a) non-rigid editing and (b) rigid editing.
In the case of non-rigid editing, current editing systems' results are not aligned with the target prompt, generating original input videos or incorrectly synthesizing original content (\eg trees) and required variations (\eg wings).
However, these models with the NeuEdit framework demonstrate effective non-rigid editing on various targets including human and object.
It is also notable that motion editing about thumbs-up is conditionally performed according to the visibility of the skier's hand, this is because the visual neutralization is sensibly applied to visible editing factors (\ie hand).
We provide further analysis of this in Section \ref{ablation}.
In the case of rigid editing (\ie top: style transfer, bottom: object overlay), the current models and the models with NeuEdit provide qualitatively proper modifications. 
But, in detail, only the models under NeuEdit framework maintain a finer fidelity to the unedited region (\ie yellow box) in the video.
It is considered that feature commonality between neutral prompt and target prompt, excluding the editing factor, improves selective fidelity in the model.
At the bottom, we also edited a man in kite-surfing to resemble Spider-Man. 
Interestingly, the model with NeuEdit also modifies the action of catching a kite as catching it with a spider web.
We consider the neutral video shown in the pink box contributes to effective editing, mitigating the restriction by the original content in the area to be edited.
%
\begin{table*}[h]
\centering
\footnotesize
\begin{center}
\begin{tabular}{lcccccccc} \toprule[1pt]
& \multicolumn{2}{c}{\bf Textual Alignment} &\multicolumn{3}{c}{\bf Fidelity to Input Video} &\multicolumn{2}{c}{\bf Frame Consistency} & {\bf Human}
\\ \cline{2-9}
& $\textrm{CLIP}^{\star}$ ↑ & PickScore ↑ & PSNR ↑ & LPIPS ↓ & SSIM ↑ & $\textrm{CLIP}^{\dagger}$ ↑ & FVD ↓  &Preference ↑ \\ \midrule
TAV \cite{wu2022tune}           &22.6 / 27.1 &19.5 / 20.2 &13.1 / 14.1 &0.1813 / 0.1934 &0.621 / 0.653 &0.921 / 0.952 &3481 / 3392 &0.14 \\
TAV + NeuEdit                &27.6 / 28.5 &20.6 / 20.9 &19.2 / 18.9 &0.1438 / 0.1411 &0.706 / 0.711 &0.962 / 0.971 &3270 / 3151 &0.86 \\ \midrule 
FateZero \cite{qi2023fatezero}  &21.2 / 26.1 &19.4 / 20.1 &14.1 / 13.6 &0.1653 / 0.1731 &0.636 / 0.643 &0.958 / 0.960 & 3319 / 3106 &0.34 \\ 
FateZero + NeuEdit           &27.3 / 28.7 &20.1 / 21.2 &16.8 / 17.3 &0.1621 / 0.1724 &0.637 / 0.657 &\textbf{0.969} / 0.968 & 3209 / 3071 &0.66 \\ \midrule
Video-P2P \cite{liu2023video}   &22.5 / 27.2 &19.6 / 20.0 &14.7 / 15.5 &0.1738 / 0.1814 &0.645 / 0.677 &0.961 / 0.958 & 3231 / 3095 &0.38 \\
Video-P2P + NeuEdit          &\textbf{27.9} / \textbf{29.6} &\textbf{20.9} / \textbf{21.3} &\textbf{19.3} / \textbf{19.8} &\textbf{0.1298} / \textbf{0.1388} &\textbf{0.727} / \textbf{0.733} &0.966 / \textbf{0.973} & \textbf{3135} /  \textbf{2953} &0.62 \\ \bottomrule[1pt]
\end{tabular}
\end{center}
\caption{Evaluations about edited videos based on DAVIS and TGVE in terms of non-rigid/rigid type editing corresponding to textual alignment, fidelity to input video, frame consistency, and human preference. $\textrm{CLIP}^{\star}$: text-video clip score,
$\textrm{CLIP}^{\dagger}$: image-image clip score.}
\label{mytab:1}
\end{table*}
\paragraph{Quantitative Results.}
%
%
Table \ref{mytab:1} presents evaluations of the non-rigid/rigid editing on videos of DAVIS and TGVE\footnote{Appendix C provides further results on UCF101\cite{soomro2012ucf101}} of recent editing systems with the NeuEdit about four assessments (\ie alignment, fidelity, consistency, human evaluation).
The effectiveness of NeuEdit is confirmed in all models. 
Especially in non-rigid editing, textual alignment is significantly improved.
Fidelity evaluates the preservation of unedited areas in video, such that we measure fidelity after masking identical regions related to editing. 
%
%
The fidelity is effectively enhanced in the tuning-based models (\ie TAV, Video-P2P) than tuning-free model (\ie FateZero), which tells that neutral prompt contributes to improving fidelity.
\begin{figure}[t]
\centering
    \includegraphics[width=1.0\linewidth]{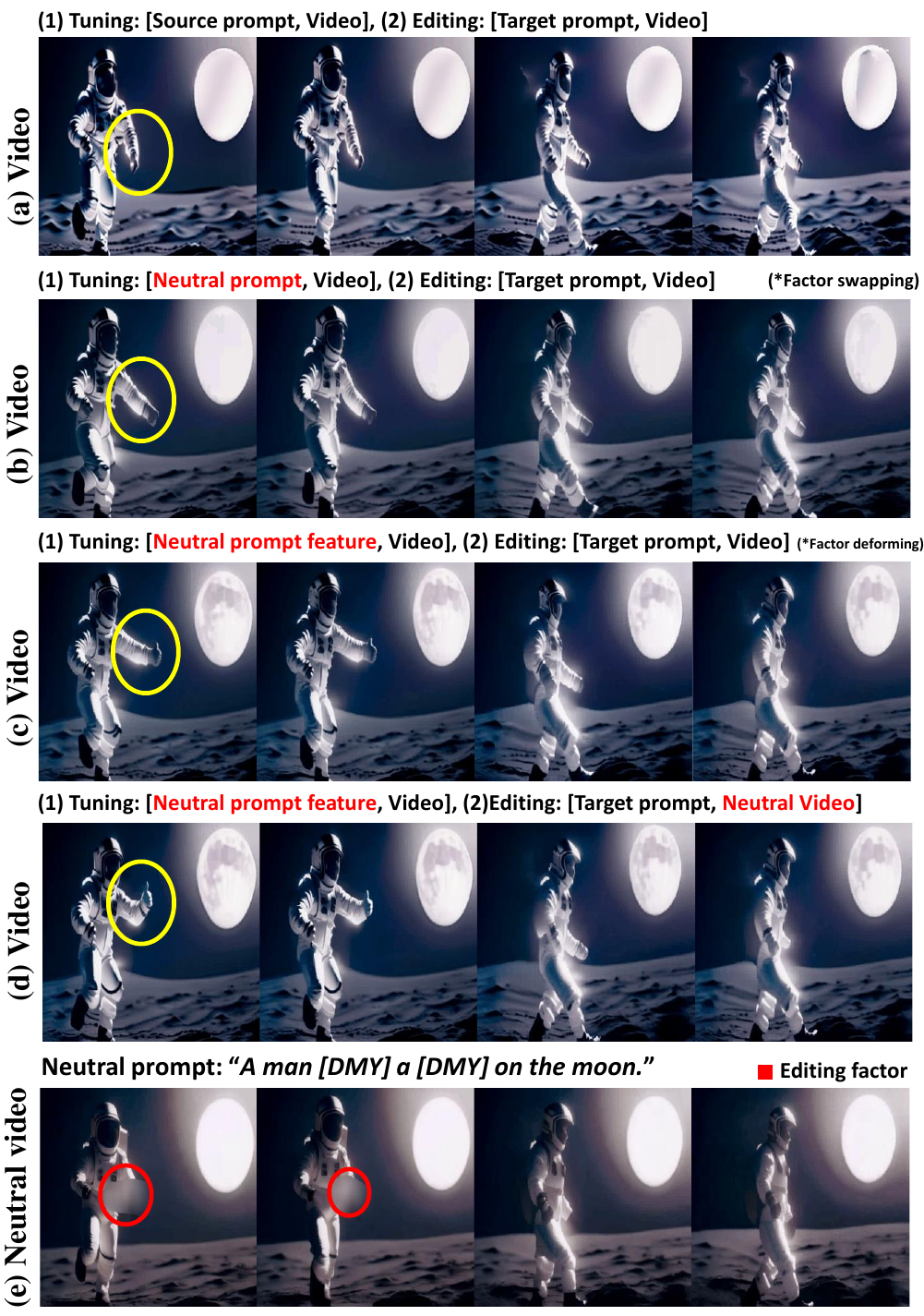}
   \caption{Ablation studies about neutral video and neutral prompt. The input video and target are shown in Figure \ref{fig:teaser}.}
\label{fig:ablation}
\end{figure}
\begin{figure}[t]
\centering
    \includegraphics[width=1.0\linewidth]{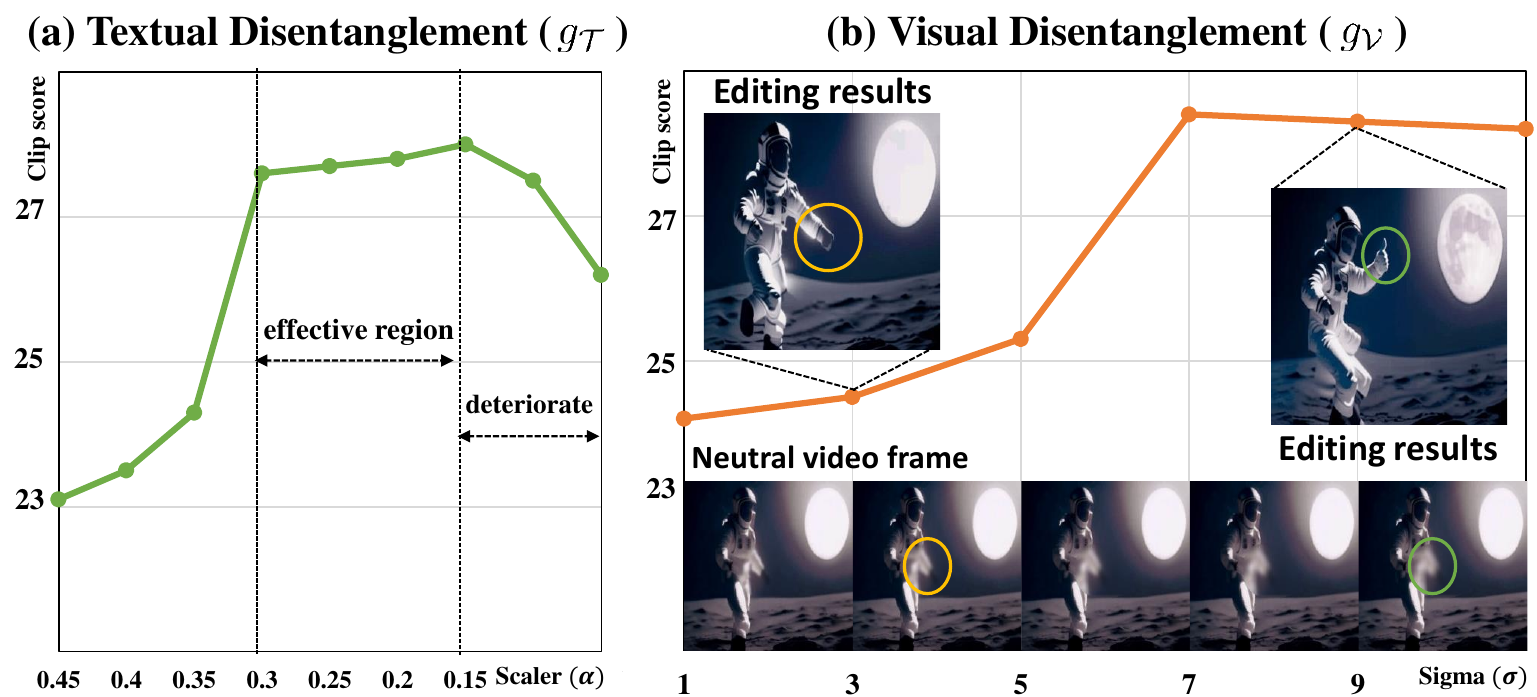}
   \caption{Sensitivity analysis of textual and visual semantic disentanglement. The $g_{\mathcal{T}}$ employs feature down-scaling with scaler $\alpha$ and $g_{\mathcal{T}}$ employs Gaussian blurring with $\sigma$. Single frame of neutral video according to the blurring by $\sigma$ is shown below the curve (b).}
\label{fig:factor_deform}
\end{figure}
\subsection{Ablation Study}
\label{ablation}
Figure \ref{fig:ablation} presents ablation studies about neutral video and neutral prompt in terms of motion editing.
In Figure \ref{fig:ablation} (a), the current model is ineffective in motion editing (\ie thumbs-up), resulting in a video closely resembling the input video.
We applied neutralization to this model, implementing its process step by step.
In (b) and (c), we show results from neutral prompt tuning applied to the editing model.
The results (b) from neutral prompt with factor swapping, while (c) stems from neutral prompt feature with factor deforming.
As shown in yellow circles in (b) and (c), they show the astronaut slightly extends his arm to give a thumbs-up, 
The action in (c) seems more effective than (b).
We consider that neutral prompt features identify differences among editing factors to be emphasized.
%
%
Nevertheless, unnatural motion editing persists due to constraints posed by the original arm motion.
Video (d) is edited results with neutral video (e), where it showcases dynamic motion edits, such as bending arms for a thumbs-up.
Notably, neutralization is sensibly applied to editing factors.
When the hand disappears behind the leg, the neutralization effect (\ie red circle) vanishes, facilitating natural motion recovery akin to temporal motion change in the result.
Figure \ref{fig:factor_deform} shows sensitivity analysis of factor deforming in textual and visual semantic disentanglement.
The textual disentanglement $g_{\mathcal{T}}$ is modulated by feature down-scaler $\alpha$. 
Neutral prompt features are effective for values below 0.3, but below 0.15, it shows deterioration, damaging the distinguishability among editing factors.
For the disentanglement $g_{\mathcal{V}}$, the Gaussian blurring is controlled by the $\sigma$.
Effective motion editing occurs for $\sigma > 3$, aligning with ambiguity caused by blurring (\ie yellow to green in Figure \ref{fig:factor_deform} (b)). In lower values, it is restricted by the original motion of video.
\section{Conclusion}
This paper proposes diffusion-based video editing framework referred to as Neutral Editing (NeuEdit), which enables complex non-rigid editing of a person/object in a video.
NeuEdit introduces a `neutralization' concept to enhance the current tuning-editing process of diffusion-based editing systems in a model-agnostic manner.
Extensive experiments validate its editability and visual effectiveness.
%



%% file: sec/X_suppl.tex
\clearpage
\setcounter{page}{1}
\maketitlesupplementary
%
%
%

\appendix

\section{Broader Impacts and Ethic Statements}
Visual generative models raise several ethical concerns such as illegal counterfeit content, potential invasion of privacy, and fairness issues.
Our work also relies on the underlying framework of these generative models, making it vulnerable to these concerns.
Therefore, effectively addressing these concerns is required, where various regulations should be prepared including technical safeguards.
Crucially, researchers should take responsibility for these concerns and actively make an effort to build technical safeguards.
Therefore, to mitigate potential concerns and hold transparency, we will release our source code including specifications of models and data that we employed under a license encouraging ethical and legal usage.
We also consider introducing further regulations such as learning-based digital forensics and digital watermarking.
Collectively, these measures aim to navigate the ethical landscape of visual generative models, fostering their responsible and beneficial use.

\section{Limitation and Future work}
\begin{figure*}[t!]
\centering
    \includegraphics[width=0.9\linewidth]{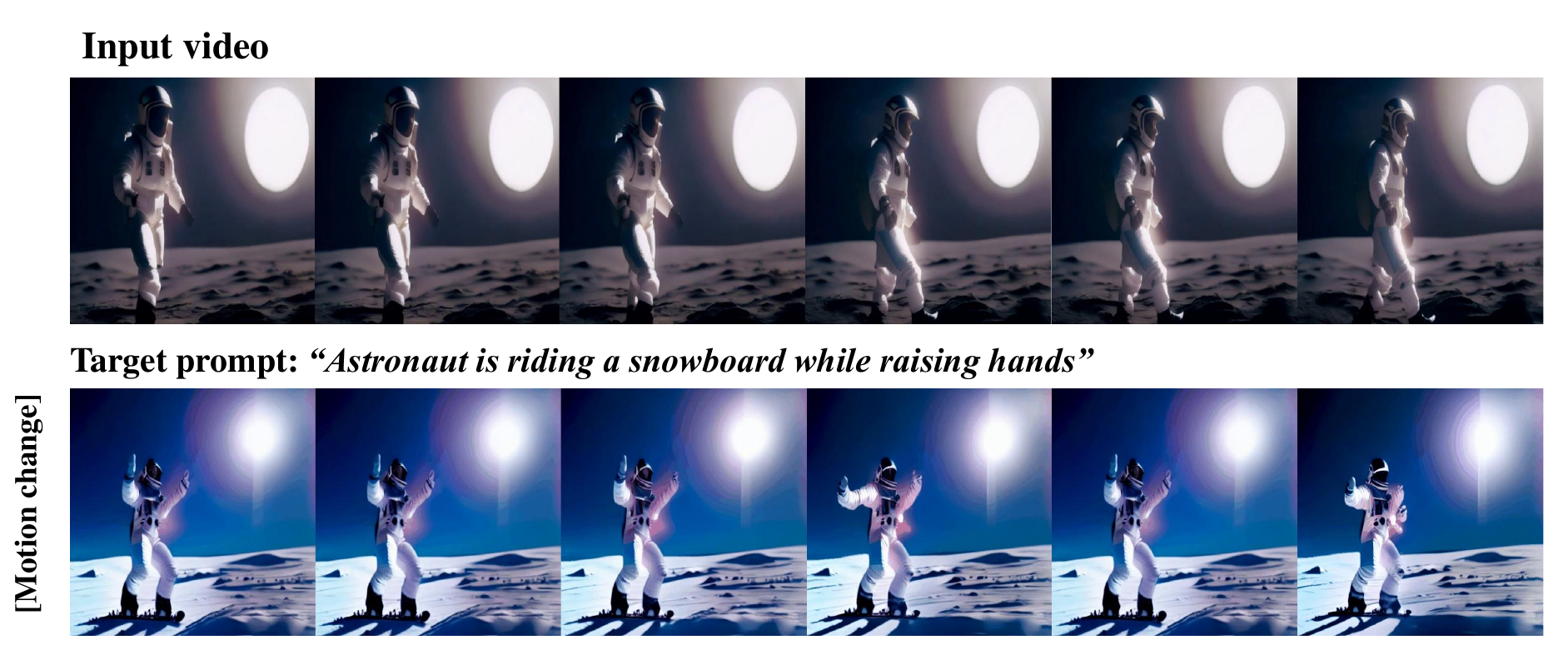}
   \caption{Failure case by editing bias. The scene is unintentionally changed into a snow background correlated with riding a snowboard.}
\label{fig:failure}
\end{figure*}

The editing systems seem to be susceptible to unintended bias in modifying required attributes.
For instance, Figure \ref{fig:failure} shows the failure case of our method.
When modifying specific attributes (e.g., motion of riding snowboard) of an object in a video, the scene in the video is also changed into a context (\eg snow) primarily associated with the desired attributes.
%
%
We define this issue as editing bias and our future work is to mitigate the editing bias.
Although NeuEdit is also successfully applicable in image non-rigid editing including still-pose editing (\ie Figure \ref{fig:image_editing}), in a video domain, it was also challenging to edit a motion of a moving object to be still.
As the temporal attention in the video diffusion model performs to preserve temporal consistency, this consistency is reflected in the object to be edited, such that a still pose is made, but follows the movements of the object.
\section{Further details and more evaluations}
We present further details of our experiments including implementations, evaluations, and results.
\subsection{Implementation details}
For video encoding, we utilize VQ-VAE \cite{van2017neural}, which provides patch-wise features of each frame, and for text encoding the CLIP model (ViT-L/14) \cite{radford2021learning} is employed.
In the visual neutralizing, we applied bicubic interpolation to restore the original scale of frames from each $i$-th frame editing factor score $\mathbf{z}_{\mathcal{V}}^{i}$.
In the case of the cross-attention module, the cross-attention weights of the first up-block attention layer in Stable Diffusion U-Net \cite{rombach2022high} is utilized, where the attention map with the size of $16 \times 16$ is constructed.
Empirically, other mid-block and up-block layers can also properly work for designing the visual editing factor.
However, the down-block layers were not effective due to insufficient early multimodal interactions between text and image.

For the details of neutral video, we set visual editing factor scores below 0.2 uniformly to zero. This approach improves the effectiveness of editing in the targeted region by establishing a clear boundary between the visual regions associated with the editing factor and those that are not.
%
%
%
%
\subsection{Evaluation details}
\begin{figure}[t]
\centering
    \includegraphics[width=1.0\linewidth]{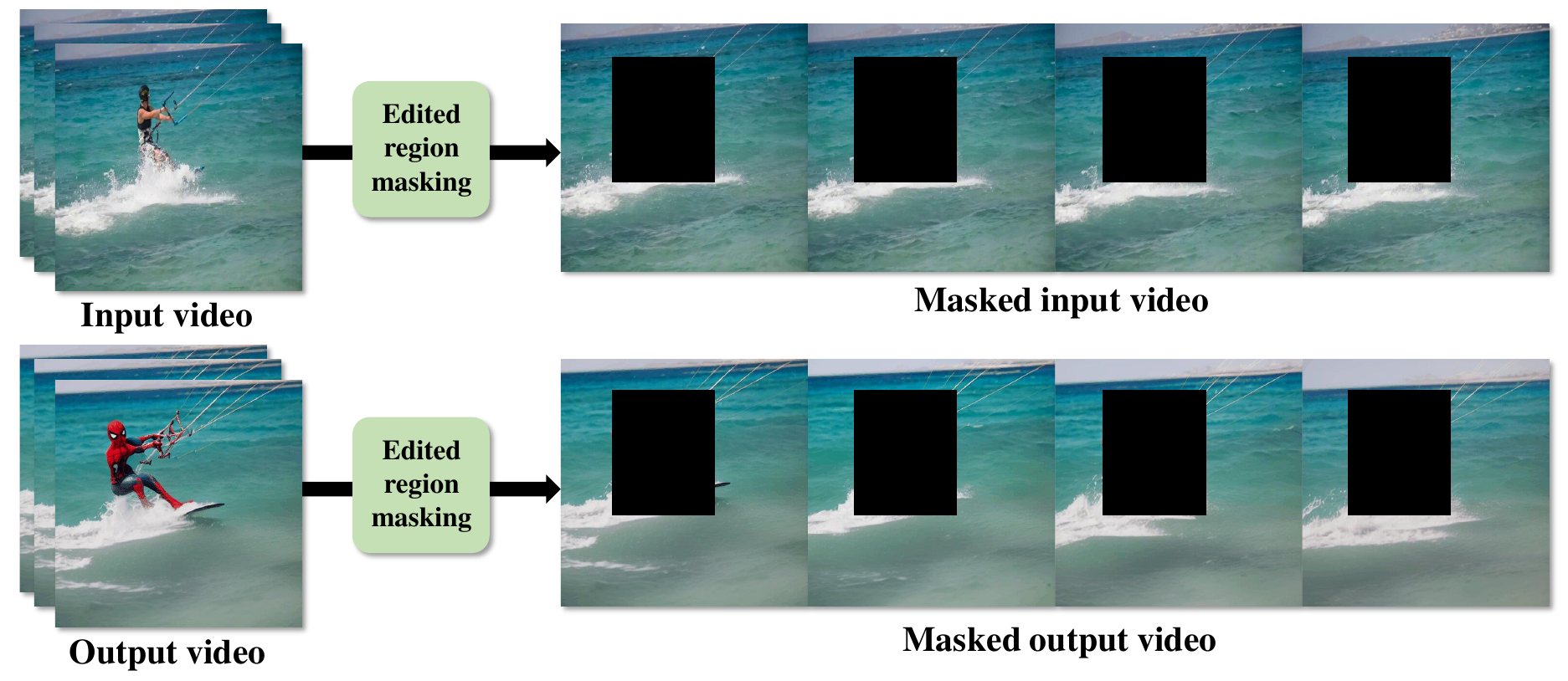}
   \caption{Illustration of masked input and output video for measuring fidelity metrics. Identical masking is applied between the input video and the output edited video.}
\label{fig:fidelity_masking}
\end{figure}

\paragraph{Fidelity evaluation.}
In order to measure fidelity to input video, as shown in Figure \ref{fig:fidelity_masking}, we applied the identical zero mask to the edited area in the input video and output video.
Applying a mask to the area to be edited allows us to measure the similarity and commonality between the input video and the output video in terms of preservation of unedited content, producing the score of PSNR, LPIPS, and SSIM.
In the early study, we attempted several automatic detectors such as segmentation-based detectors \cite{kirillov2023segment} to identify areas to be edited in input and output video, but the specifications by humans were the most accurate.
%
%
%
%
%
\paragraph{Human evaluation.}
Human evaluation is performed to measure the preference for edited results according to a given target prompt.
We conducted a survey about discrete selection between the outcomes of current editing systems and those generated using the NeuEdit framework.
A survey involving 36 participants was conducted, incorporating diverse academic backgrounds (\eg engineering, literature, art) and including those who speak English as their native language and those who do not.
%
%
\subsection{Evaluations on videos of different domains}
\label{eval:dataset_ucf}
\paragraph{UCF101} Collected from YouTube, UCF101 \cite{soomro2012ucf101} dataset is designed for action recognition, comprising 101 action categories.
To measure the video editing performances according to the editing model with NeuEdit framework and without the framework, We selected 83 videos from the dataset.
Overall, performances of all models (\ie TAV: Tune-A-Video, FateZero, and Video-P2P) using NeuEdit framework are enhanced in terms of textual alignment, fidelity, and consistency.
Furthermore, the tuning-editing models (\ie TAV, Video-P2P) of NeuEdit framework exhibited a significant enhancement in fidelity, similar to the DAVIS dataset, demonstrating general effectiveness in videos.
%
%
\begin{table*}[h]
\centering
\footnotesize
\begin{center}
\begin{tabular}{lcccccccc} \toprule[1pt]
& \multicolumn{2}{c}{\bf Textual Alignment} &\multicolumn{3}{c}{\bf Fidelity to Input Video} &\multicolumn{2}{c}{\bf Frame Consistency} & {\bf Human}
\\ \cline{2-9}
& $\textrm{CLIP}^{\star}$ ↑ & PickScore ↑ & PSNR ↑ & LPIPS ↓ & SSIM ↑ & $\textrm{CLIP}^{\dagger}$ ↑ & FVD ↓  &Preference ↑ \\ \midrule
TAV \cite{wu2022tune}           &22.3 / 26.7 &19.2 / 19.7 &12.7 / 13.8 &0.1911 / 0.2031 &0.573 / 0.609 &0.913 / 0.937 &3537 / 3441 &0.15 \\
TAV + NeuEdit                &26.8 / 27.9 &20.1 / 20.6 &18.2 / 17.4 &0.1532 / 0.1509 &0.658 / 0.669 &0.946 / 0.952 &3321 / 3202 &0.85 \\ \midrule 
FateZero \cite{qi2023fatezero}  &21.0 / 25.7 &19.1 / 19.6 &13.1 / 13.2 &0.1749 / 0.1836 &0.583 / 0.598 &0.936 / 0.941 & 3372 / 3151 &0.30 \\ 
FateZero + NeuEdit           &26.7 / 27.9 &20.0 / 20.8 &15.9 / 16.9 &0.1723 / 0.1811 &0.589 / 0.601 &0.947 / 0.946 & 3252 / 3123 &0.70 \\ \midrule
Video-P2P \cite{liu2023video}   &22.3 / 26.8 &19.3 / 19.6 &14.2 / 14.9 &0.1821 / 0.1923 &0.598 / 0.623 &0.942 / 0.939 & 3282 / 3142 &0.31 \\
Video-P2P + NeuEdit          &\textbf{27.1} / \textbf{28.4} &\textbf{20.5} / \textbf{21.1} &\textbf{18.4} / \textbf{18.9} &\textbf{0.1312} / \textbf{0.1413} &\textbf{0.676} / \textbf{0.681} &\textbf{0.948} / \textbf{0.954} & \textbf{3182} /  \textbf{3001} &0.69 \\ \bottomrule[1pt]
\end{tabular}
\end{center}
\caption{Evaluations about edited videos based on UCF101 in terms of non-rigid/rigid type editing corresponding to textual alignment, fidelity to input video, frame consistency, and human preference. $\textrm{CLIP}^{\star}$: text-video clip score,
$\textrm{CLIP}^{\dagger}$: image-image clip score.}
\label{mytab:ucf}
\end{table*}
\section{Proof for the closed form of KL divergence in reverse diffusion process}
The reverse process of denoising diffusion probabilistic models is to approximate $q(x_{t-1}|x_{t})$ using parameterized Gaussian transitions $p_{\theta}(x_{t-1}|x_{t}) = \mathcal{N}(x_{t-1};\mu_{\theta}(x_t,t),\sigma_{\theta}(x_{t},t))$. 
Considering whole $T$ step parameterized transitions, these are sequentially constructed as given below:
\begin{equation}
p_{\theta}(X) = p_{\theta}(x_{T})\prod_{t=1}^{T}p_{\theta}(x_{t-1}|x_{t}),
\end{equation}
where we take $X = x_{0:T}$ and it starts at normal distribution $p(x_{T}) = \mathcal{N}(x_{T};0,\textit{I})$.
To optimize the $p_{\theta}(X)$, training objective is to maximize log-likelihood $\textrm{log}(p_{\theta}(X))$, where we can also apply variational inference by maximizing the variational lower bound $-L_{VLB}$ as given below:
\begin{equation}
\begin{aligned}
-L_{VLB} = \textrm{log}p_{\theta}(X) - D_{\textrm{KL}}(q(Z|X)||p_{\theta}(Z|X))\\
\leq \textrm{log}p_{\theta}(X),
\end{aligned}
\end{equation}
where $D_{\textrm{KL}}$ is the Kullback-Leibler divergence (KL divergence) and the $Z$ is latent variable by reparametrization trick used in the variational auto-encoder.
The $q$ can be any distribution that we can address with ease.
We leverage this inequality condition as $-\textrm{log}p_{\theta}(X) \leq L_{VLB}$. 
The $L_{VLB}$ can be expanded out as $L_{VLB} = L_{T} + L_{T-1} + \cdots + L_{0}$, where they are defined with $1 \leq t \leq T$ as given below:
\begin{equation}
\begin{aligned}
&L_{T} = D_{\textrm{KL}}(q(x_{T}|x_{0})||p_{\theta}(x_{T})),\\ 
&L_{t} = D_{\textrm{KL}}(q(x_{t}|x_{t+1},x_{0})||p_{\theta}(x_{t}|x_{t+1})), \\ 
&L_{0} = -\textrm{log}p_{\theta}(x_{0}|x_{1}).
\end{aligned}
\end{equation}
Therefore the terms about $L_{t}$ make the closed form of KL divergence under step $t$ with a range of $0 \leq t \leq T$. 
\section{DDIM sampling and DDIM inversion}
To accelerate the reverse process of DDPM, denoising diffusion implicit model (DDIM) \citep{song2020denoising} is proposed, it samples latent features with a small number of denoising steps as: 
\begin{equation}
\begin{aligned}
z_{t-1} = \sqrt{\frac{\alpha_{t-1}}{\alpha_{t}}}z_{t} + \left( \sqrt{\frac{1}{\alpha_{t-1}}-1} - \sqrt{\frac{1}{\alpha_{t}}-1} \right) \epsilon
\end{aligned}
\end{equation}
We can also reverse this process to make latent noise again, which gives corresponding latent features as below: 
\begin{equation}
z_{t+1} = \sqrt{\frac{\alpha_{t+1}}{\alpha_{t}}}z_{t} + \left( \sqrt{\frac{1}{\alpha_{t+1}} - 1} - \sqrt{\frac{1}{\alpha_{t}} - 1} \right) \epsilon,
\end{equation}
where it is referred to as DDIM inversion process, which maintains higher fidelity to the input than just initially starting from Gaussian noise.

\begin{figure}[t]
\centering
    \includegraphics[width=1.0\linewidth]{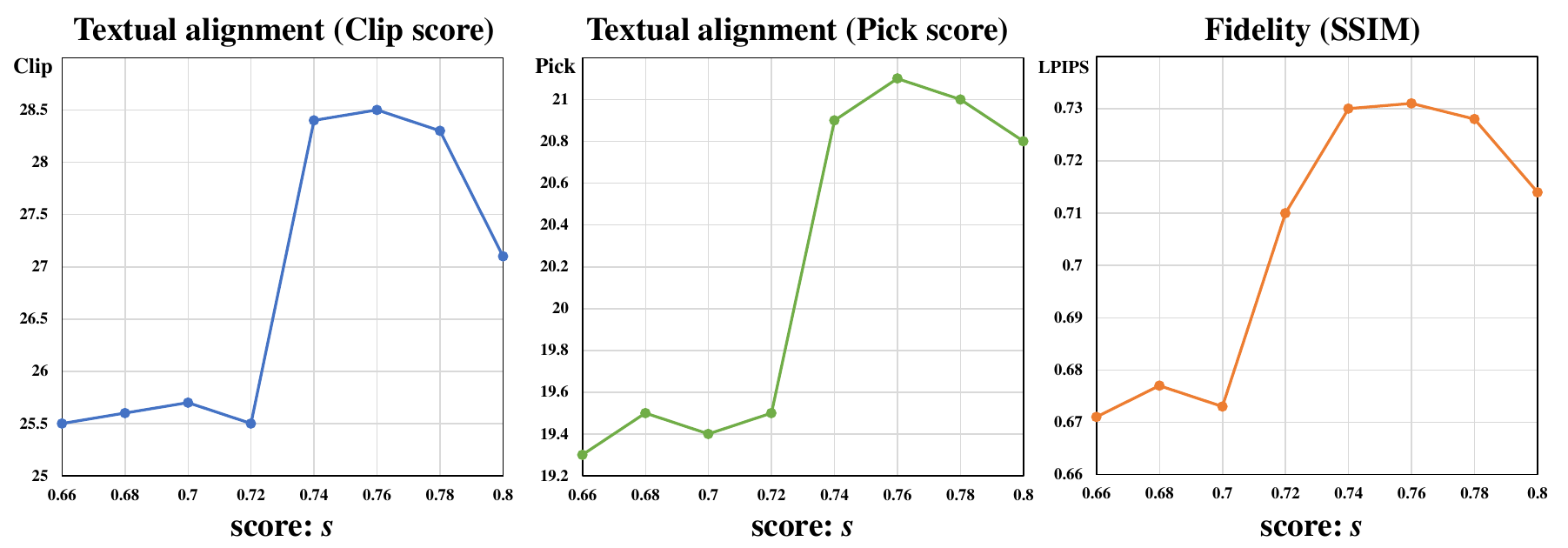}
   \caption{Sensitivity analysis of the score $s$ for selecting editing factor for factor swapping in terms of textual alignment (\ie ClIP score, Pick score) and fidelity (\ie SSIM).}
\label{fig:sensitivity}
\end{figure}
\section{Studies of textual semantic disentanglement}
\subsection{Sensitivity analysis of factor swapping}
Factor swapping involves the binary deletion of a word according to a decision on the editing factor. Hence, the proper selection of editing factors contributes to an efficient neutral prompt. 
To explore this aspect further, we aim to investigate the optimal operational range of factor swapping by leveraging the threshold score $s$ for deciding the editing factor. 
Figure \ref{fig:sensitivity} depicts the editing performance variations with changes in the score $s$, where the $0.74 < s < 0.78$ presents the optimal range for effectively working as a neutral prompt with factor swapping.
%
%
%
%
%
%
\subsection{Ablation studies about factor deforming}
We introduce further empirical methods performed for textual semantic disentanglement. In the main paper, we employ factor deforming $\mathbf{w}_{n} = \mathbf{z}_{\mathcal{T}} \circ h(\mathbf{w}) + (1 - \mathbf{z}_{\mathcal{T}}) \circ \mathbf{w}$ with the deformable operation of down-scaling as below: 
\begin{equation}
h(\mathbf{w}) = \alpha \mathbf{w} \in \mathbb{R}^{M \times d},
\end{equation}
where the $0 \leq \alpha < 1$ is the down-scaler.
Our studies also consider other deformable operations defined as (1) deformable swapping and (2) factor blurring. The detailed explanations are presented in the following.
\paragraph{Deformable swapping}
The deformable swapping is an extended version of factor swapping.
Factor swapping involves transforming all words identified as editing factors into a unified token as a dummy token $<$DMY$>$.
The limitation of this approach is that all modified dummy tokens become indistinguishable from one another.
Therefore, we integrated the format of factor deforming with the factor swapping and adaptively changed the magnitudes of the dummy token feature according to the editing factor score as given below:
\begin{equation}
\mathbf{w}_{n} = \mathbf{z}_{\mathcal{T}} \circ \mathbf{w}_{n}^{\textrm{swp}} + (1 - \mathbf{z}_{\mathcal{T}}) \circ \mathbf{w},
\end{equation}
where $\mathbf{w}_{n}^{\textrm{swp}}$ is text feature obtained by factor swapping.
This format is available to impart distinguishable influence to dummy token features based on the editing factor score. 
\paragraph{Factor blurring}
Another attempt is factor blurring.
Similar to the factor deforming in the visual textual semantic disentanglement, we apply to blur the features related to editing factors.
Hence, rather than employing feature down-scaling, we adopt an alternative approach by introducing an additional $d$-dimensional noise feature to deform the target prompt features as given below
\begin{equation}
\mathbf{w}_{n} = \mathbf{z}_{\mathcal{T}} \circ (\mathbf{w} + \epsilon) + (1 - \mathbf{z}_{\mathcal{T}}) \circ \mathbf{w},
\end{equation}
where $\epsilon \sim \mathcal{N}(0,1)$ is the noise features added to target prompt features.
It deforms the prompt feature corresponding to editing factors.
\begin{table}[h]
\centering
\footnotesize
\begin{center}
\begin{tabular}{lccc} \toprule[1pt]
& \multicolumn{1}{c}{\bf Alignment} &\multicolumn{1}{c}{\bf Fidelity} &\multicolumn{1}{c}{\bf Consistency}
\\ \cline{2-4}
                            & $\textrm{CLIP}^{\star}$ ↑  & SSIM ↑  & $\textrm{CLIP}^{\dagger}$ ↑   \\ \midrule
factor deforming (M)            & 28.6                       & 0.730   & 0.969                  \\ \midrule
factor blurring  (A)           & 27.7                       & 0.697   & 0.948                 \\ \midrule
factor swapping (M)            & 27.5                       & 0.707   &0.951  \\ \midrule
deformable swapping (A)        & 28.1                       & 0.711   & 0.961  \\ \bottomrule[1pt]
\end{tabular}
\end{center}
\caption{Ablation study of different methods of factor deforming for textual semantic disentanglement. M: method in Main paper, A: method is Appendix}
\label{mytab:factor_deforming_ablation}
\end{table}

Table \ref{mytab:factor_deforming_ablation} presents ablation studies in terms of factor deforming used for textual semantic disentangling.
The first section is the original factor deforming with feature-down scaling in the main paper.
The second section is the factor blurring method, which was which was less effective than other methods.
We consider that although the blurring method is effective in visual semantic disentanglement, in the case of words, it seems difficult to obtain meaningful information from the blurred features that the network understands due to words' discrete characteristics.
The third and fourth sections are factor swapping.
Building distinguishable features among dummy token features enhances the editing performances in deformable factor swapping.
This demonstrates that the editing factor scores properly contain information about how effective each factor is in performing the editing.
\begin{figure}[h!]
\centering
    \includegraphics[width=1.0\linewidth]{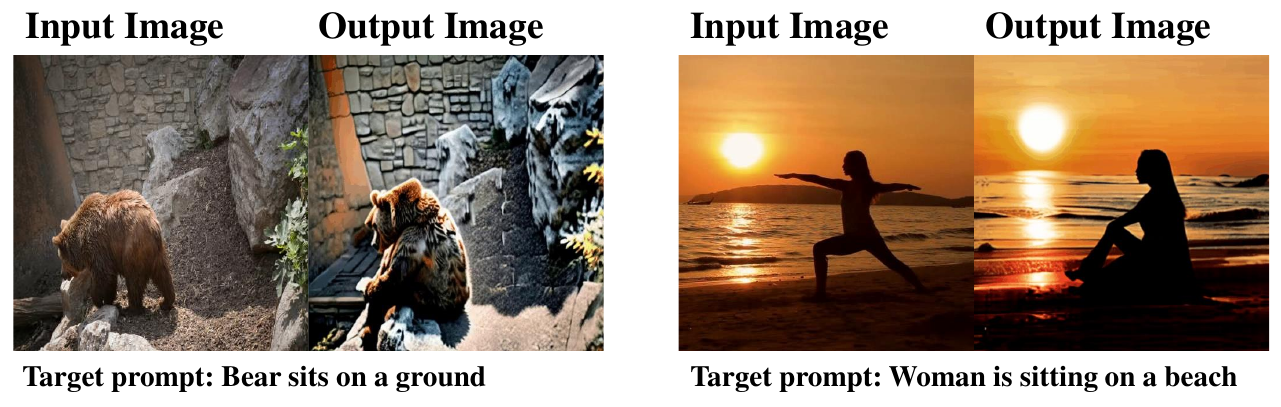}
   \caption{Application of NeuEdit into image editing about non-rigid editing (\eg pose variations).}
\label{fig:image_editing}
\end{figure}
\section{Further qualitative results}
\paragraph{Application to image editing}
The NeuEdit framework can be structurally applied to any diffusion-based editing system, such that we extend our work in image editing.
Figure \ref{fig:image_editing} shows image editing results under the NeuEdit framework.
It is notable that the non-rigid editing is also successfully applied based on the input image.
%
%
\paragraph{Further qualitative results}
In the following, we showcase our qualitative editing results in terms of non-rigid and rigid editing.
All the videos in our experiments are sourced from publicly available sources in \cite{pont20172017,soomro2012ucf101,esser2023structure,wu2022tune}.
\newpage
\begin{figure*}[t!]
\centering
    \includegraphics[width=0.85\linewidth]{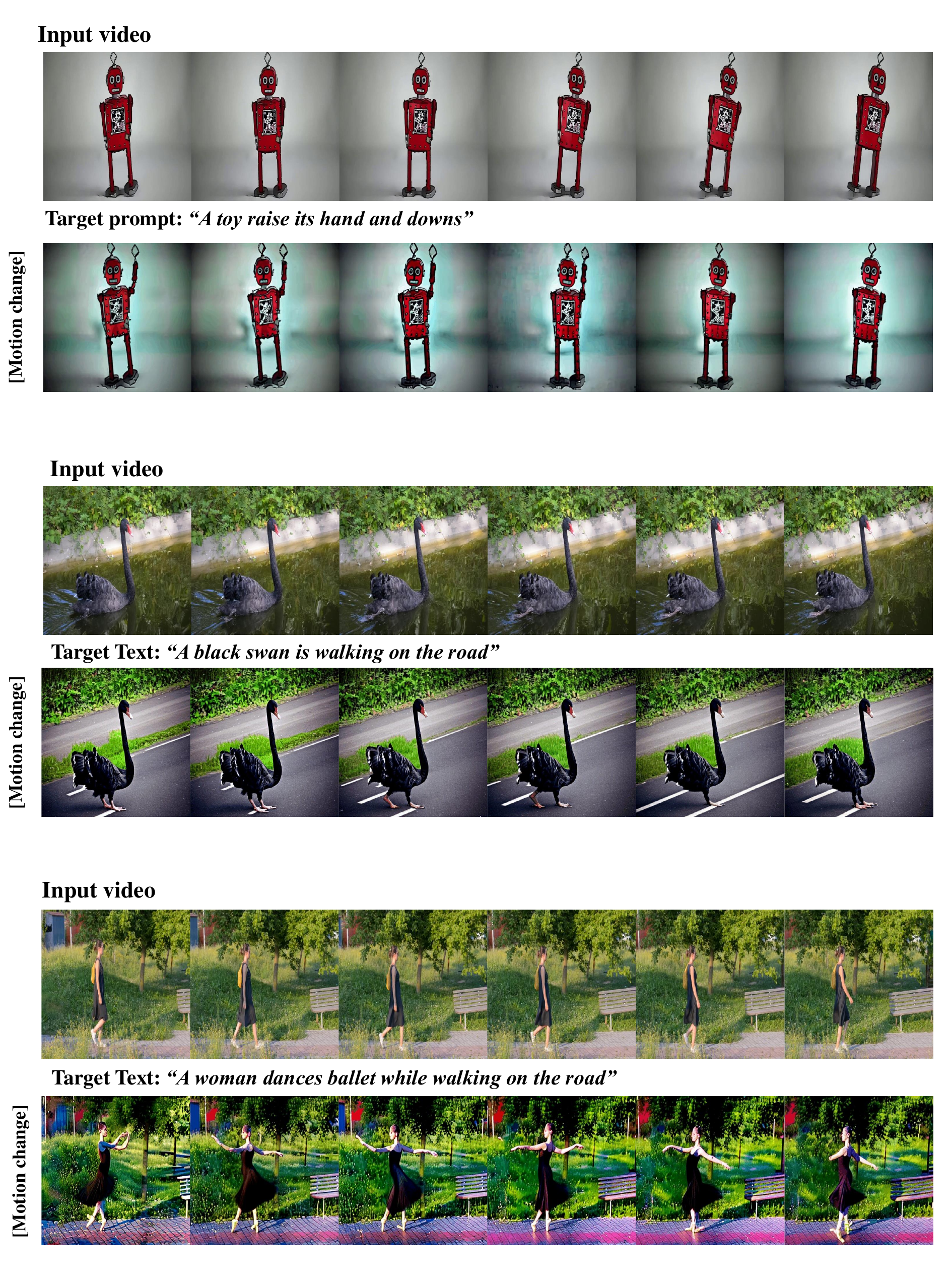}
   \caption{Illustration non-rigid editing including motion change.}
\label{fig:motion}
\end{figure*}
\newpage
\begin{figure*}[t!]
\centering
    \includegraphics[width=0.9\linewidth]{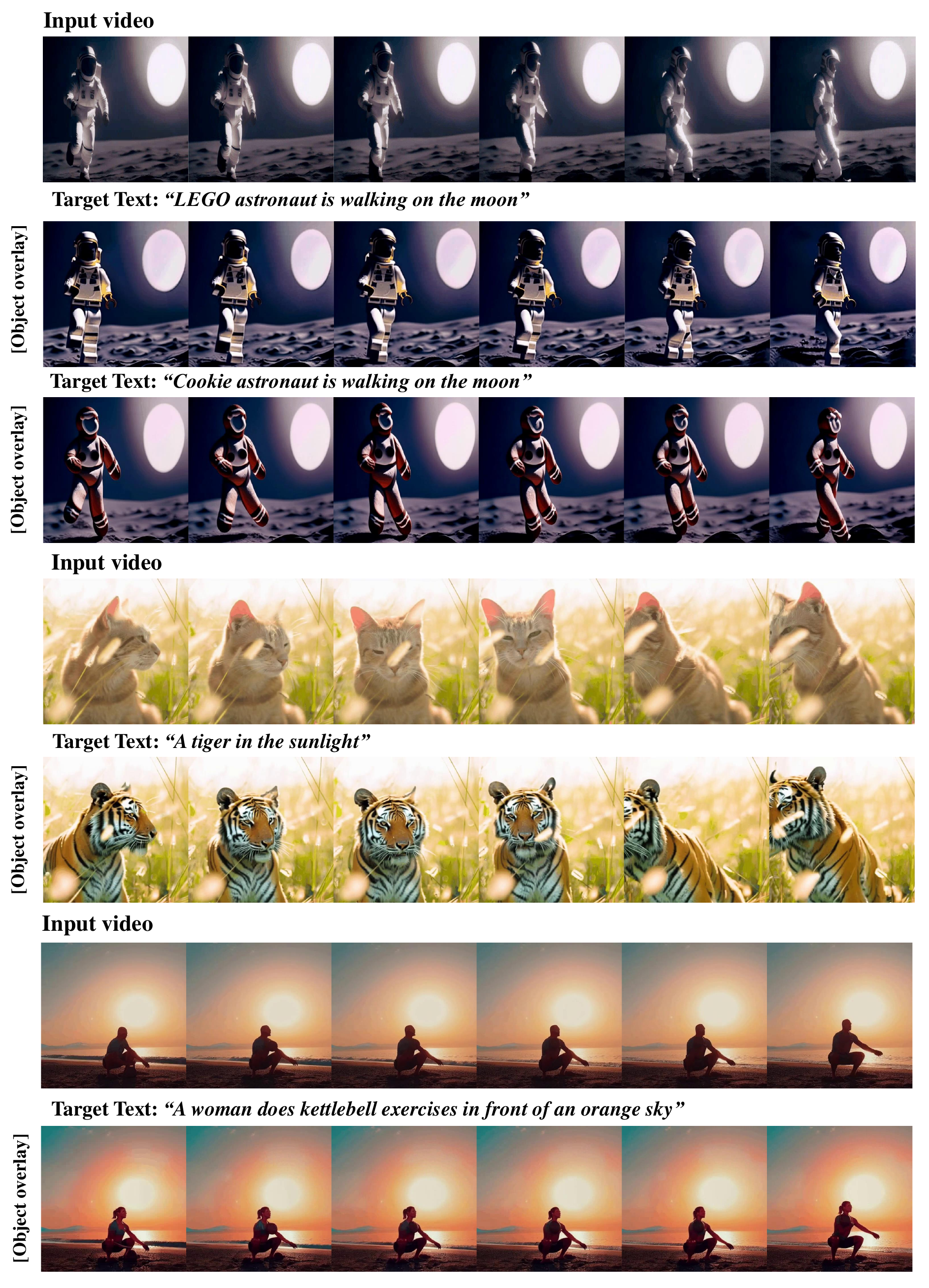}
   \caption{Illustration of rigid editing about object overlay.}
\label{fig:overlay}
\end{figure*}
\newpage
\begin{figure*}[t!]
\centering
    \includegraphics[width=0.9\linewidth]{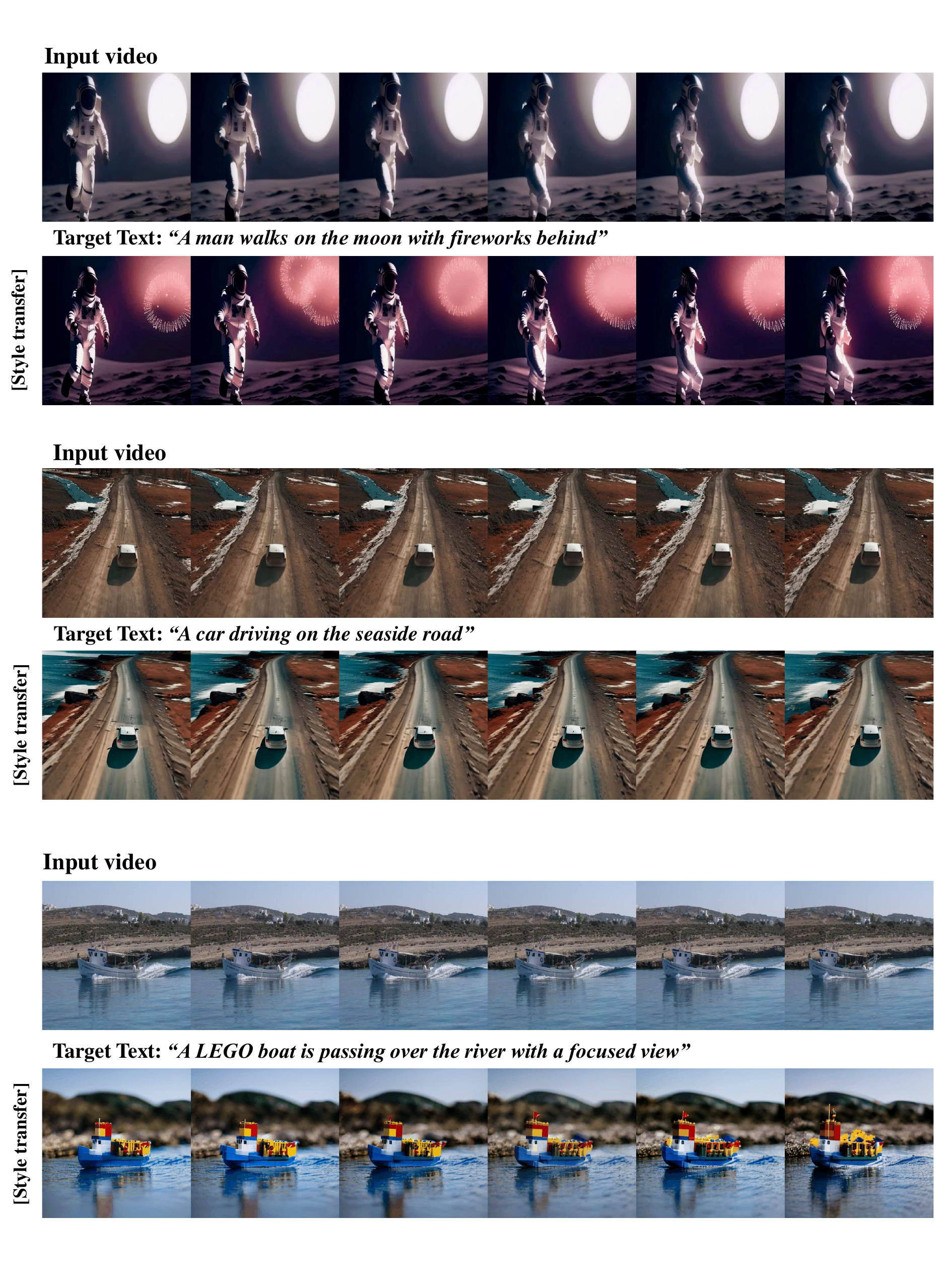}
   \caption{Illustration of rigid editing about style transfer.}
\label{fig:style}
\end{figure*}
\newpage
